# Pump It Up: Predict Water Pump Status using Attentive Tabular Learning

Karan Pathak

Computer Science Dept.

Vellore Institute of Technology, Vellore

karan.pathak2013@vitalum.ac.in

L Shalini

Computer Science Dept.

Vellore Institute of Technology, Vellore

lshalini@vit.ac.in

**ABSTRACT**

Water crisis is a crucial concern around the globe. Appropriate and timely maintenance of water pumps in drought-hit countries is vital for communities relying on the well. In this paper, we analyze and apply a sequential attentive deep neural architecture, TabNet, for predicting water pump repair status in Tanzania. The model combines the valuable benefits of tree-based algorithms and neural networks, enabling end-to-end training, model interpretability, sparse feature selection, and efficient learning on tabular data. Finally, we compare the performance of TabNet with popular gradient tree-boosting algorithms like XGBoost, LightGBM, CatBoost, and demonstrate how we can further uplift the performance by choosing focal loss as the objective function while training on imbalanced data.

**1. INTRODUCTION**

Water is one of the essential resources for a human being for its survival. Even after advances in water cleaning technology, millions of people in undeveloped and developing nations still rely on direct groundwater supply through wells. The governments have installed electrical pumps for pumping out groundwater, but these often need repair work, which could potentially disrupt the water supply of communities residing around the well. In Tanzania, a total of $1.42 billion has been donated for fixing the water crises by constructing water pumps [2], but a smart understanding of these pump's failure and timely maintenance remains a big challenge.

In this paper, we use TabNet[5], a deep neural network (DNN) designed for tabular data to efficiently predict water pumps that need repair. We use the data provided by the Tanzanian government and Taarifa, an open platform, for classifying the existing water points into three categories – functional, functional needs repair, and non-functional [1, 15].

We experimented with different tree-based gradient boosting algorithms [3,4,6] and sequential attention based DNN, TabNet [5]. TabNet combines the key benefits of tree-based algorithms (feature selection and interpretability) and DNNs (end-to-end learning). Unlike tree-based ensembles like Random Forest [10], XGBoost [4] and CatBoost [3], TabNet does not require any feature engineering. TabNet is trained using gradient descent-based optimization to learn different representations in the data. At each decision step, TabNet uses sequential attention for instance-wise feature selection, which helps in better learning and enables interpretability. In the results section, we present a comparison between different methodologies and TabNet.

Due to class imbalance in the dataset [15], we experimented with an α-balanced variant of the focal loss [11] as the objective function while training TabNet apart from the traditionally used categorical cross-entropy loss in a typical multi-class classification problem. By using focal loss, we gained a 1.2% increase in the overall accuracy. The experiment uses the F1 score along with the accuracy score as the evaluation metrics due to class imbalance in the dataset.

## 2. RELATED WORK

### A. Tree-based learning

Tabular data are commonly modeled using tree-based algorithms and are dominated with recent variants of ensemble decision trees [12], which increase the overall performance by reducing the model variance. Tree-based ensemble algorithms are interpretable, robust against overfitting [13], robust against outliers and noises [14], do not require scaled inputs [14], and can easily handle mixed data types.

Random Forest [10], a bagging tree-based technique, trains each of its weak tree learners parallelly, choosing random subsets of data with replacement along with random features. Previous experiments done using random forest [7, 9] on the Tanzanian water pump dataset [15] have achieved an accuracy of 77%.

The commonly used gradient boosting decision tree [16] (GBDT) algorithms are XGBoost [4], LightGBM [6] and CatBoost [3]. LightGBM combines two novel techniques [6]: Gradient-based One-Side Sampling (GOSS) and Exclusive Feature Bundling (EFB). GOSS is an instance sampling method for GBDT, which keeps all the large gradient instances, i.e., under-trained instances, and randomly samples from small gradient instances at every iteration. EFB is used to effectively reduce the number of features in a sparse feature space. CatBoost uses ordered boosting, a permutation-driven alternative to the standard gradient boosting algorithm, and an innovative categorical feature processing algorithm that converts the categorical data into appropriate representation internally without the need for any external pre-processing. CatBoost also solves the problem of prediction shift, a special kind of target leakage present in gradient boosting algorithms.

### B. DNN based learning

DNNs have proven to be more successful than any other technique for modeling and learning representation of different types of data: images [20, 21, 22], audio [26, 27, 28] and text [23, 24, 25]. Moreover, unlike tree-based learning, DNNs provide end-to-end single model learning and eliminate the need for feature engineering. Standalone fully connected neural network [8] and deep jointly-informed neural networks where DNNs are warm started with representations pre-learned by decision tree [29] for tabular data modeling have been proposed. But these yield inefficient learning. In a recent work, a DNN model combining the benefits of tree-based models and DNN is proposed in [5]. It provides end-to-end learning along with inbuilt sparse feature selection and model interpretability. It has proven to surpass the prediction accuracy of tree-based algorithms on different standard datasets.

### C. Focal loss

The focal loss was introduced [11] for the object detection problems. However, due to its effectiveness in handling imbalanced classes, it has been widely adopted in other fields of computer vision [18, 19], particularly in image classification problems [17]. The focal loss is designed to down-weight the easy examples by reducing their contribution to the overall objective function and focus on the hard negatives by putting more weight on the misclassified data points. The α-balanced variant [11] of the focal loss is similar to the balanced cross-entropy, where it puts inverse weights on class frequency.

## 3. METHODOLOGY

### A. Dataset

The water pump dataset [15] is provided by the Taarifa organization and the Tanzanian Ministry of Water. The dataset includes information of 59,400 wells, and each well is described using 39 features, out of which 34 are categorical variables, and 5 are continuous variables. The dataset has skewed labels, Fig. 1(a), as 54.3% of the wells are `functional`, 38.43% of the wells are `non-functional` and the remaining 7.27% of the wells are `functional but needs repair`. Fig. 1(b) shows a detailed bivariate analysis of continuous variables.

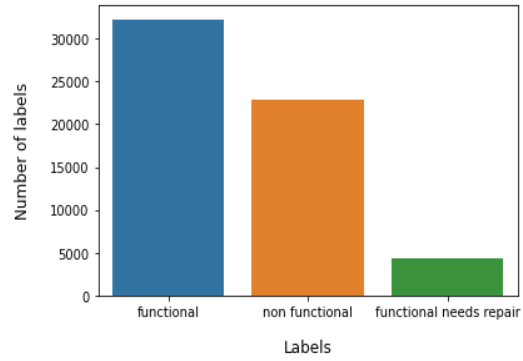

(a) Label distribution

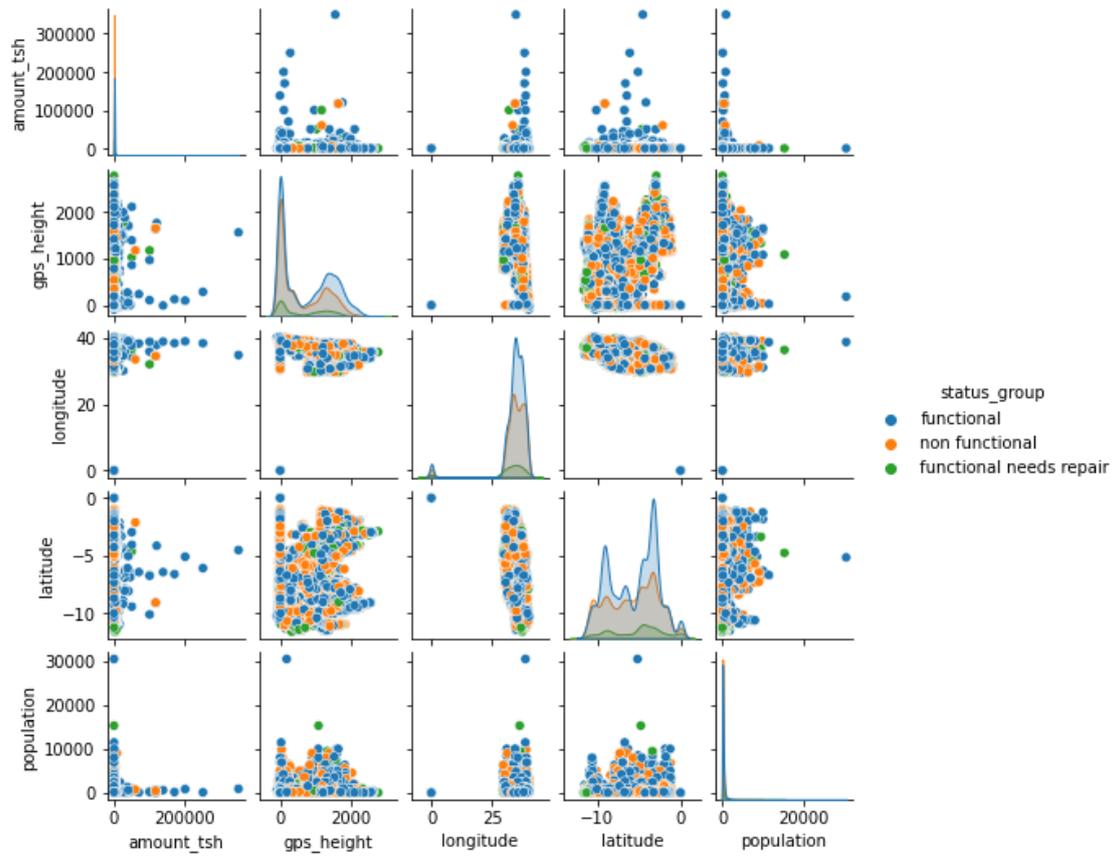

(b) Bivariate analysis of continuous variables

**Figure 1:** EDA of the water pump dataset

### B. Focal loss

Categorical cross-entropy loss tends to work fine for several different computer vision problems, but when the data labels are skewed, few of the other alternatives like the balanced cross-entropy loss and focal loss outperform the typical cross-entropy loss.

*Categorical cross-entropy (CE) loss for binary classification is defined as:*

$$p_t = \begin{cases} p & if\ y = 1 \\ 1 - p & otherwise \end{cases} \quad (1)$$

$$CE(p, y) = CE(p_t) = -\log(p_t)$$

Here, y specifies the ground-truth class and p ∈ [0, 1] is the model's estimated probability for the class with label y = 1.

A **balanced categorical cross-entropy** loss, eq. 2, can be used for the imbalanced datasets by adding a factor α, a class label weight, in the categorical cross-entropy in eq. 1.

$$CE(p, y) = CE(p_t) = -\alpha_t \log(p_t) \quad (2)$$

The balanced cross-entropy function handles the imbalance in the labels using a weighted average approach (α-factor). However, it fails to differentiate the individual contributions from the easy examples (correctly classified instances) and hard examples (incorrectly classified instances) in the loss output.

The **focal loss** addresses this issue by adding a modulating factor $(1 - p_t)^\gamma$ to the balanced cross entropy loss eq. 2, which improves the loss in a skewed label dataset. An α-balanced variant of the focal loss:

$$FL(p_t) = -\alpha_t (1 - p_t)^\gamma \log(p_t) \quad (3)$$

The modulation factor helps by focusing on training on hard negatives instead of the already correctly classified examples.

**C. TabNet**

TabNet modifies the conventional DNN architecture to reap the benefits of feature selection and interpretability in tree-based models. It provides better learning capacity than tree-based methods due to non-linearity via activation functions in the architecture. TabNet primarily consists of a feature transformer and an attentive transformer.

The **attentive transformer block,** Fig, 2(a), utilizes the information about each feature's involvement, aggregated from the previous decision steps, for selecting the most salient features for the current decision step. The attentive transformer is implemented as a learnable mask M[i] ∈ $\mathbf{R}^{B \times D}$, where B is the batch size and D is the feature dimension.

$$M[i] = sparsemax(P[i-1] \cdot hi(a[i-1])) \quad (4)$$

Sparsemax is the sparmax normalization [30]. P[i] is a cumulative function denoting each feature's contribution till step i. $h_i$ is a trainable function implemented using a fully connected layer follower by a batch normalization layer.

A **Feature transformer** is used to process the features. It consists of shared and decision step-dependent layers. The shared layer shares the same features across all the decision steps. In the paper [5], for efficient learning, it is recommended that the feature transformer block should comprise of two shared layers and two decision step-dependent layers as shown in the Fig. 2(b).

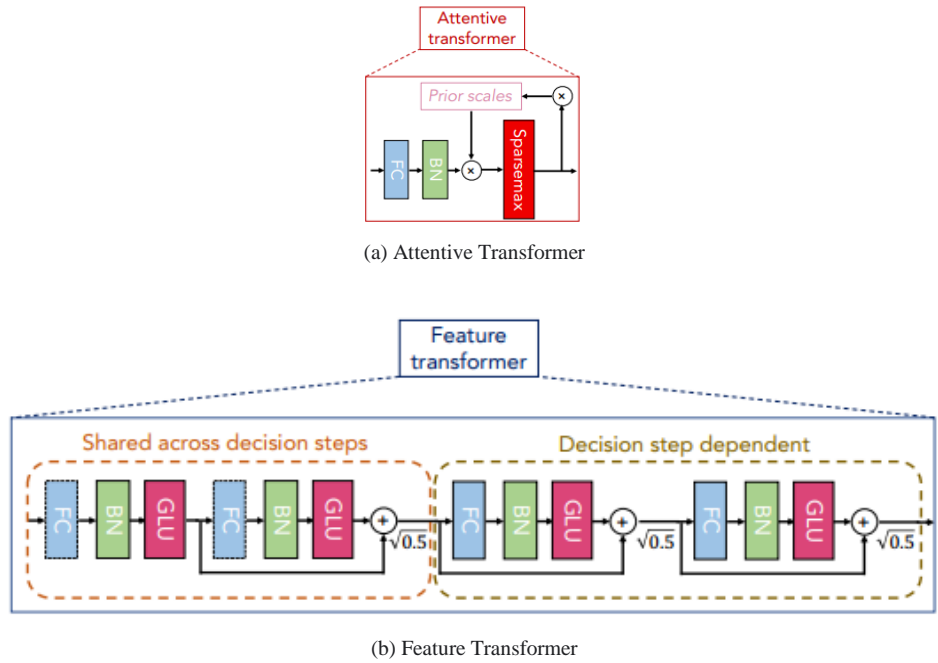

(a) Attentive Transformer

(b) Feature Transformer

**Figure 2:** Primary components of TabNet

**D. Implementation**

During the exploratory data analysis, we discover seven categorical columns containing missing data. We dropped two columns of them as more than half of the data in these columns were missing values and applied the most frequent replacement strategy in the remaining columns as these had skewed categories. We encoded the categorical variables using a label encoder.

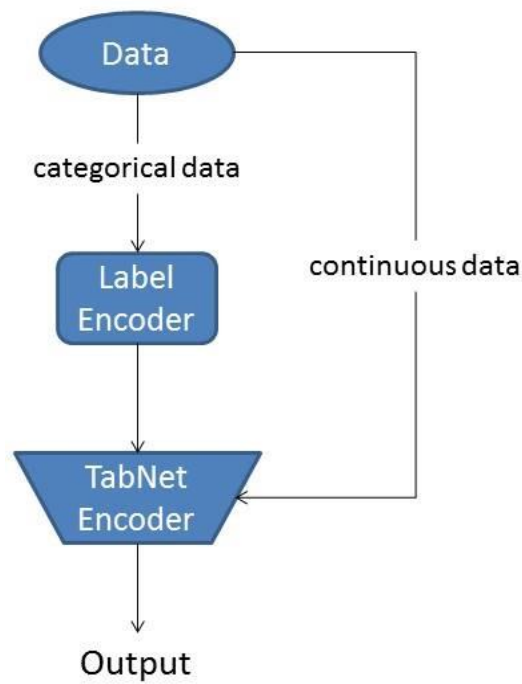

**Figure 3:** Implementation flowchart using TabNet

We experimented with TabNet (Fig. 3) and four tree-based models: XGBoost, LightGBM, CatBoost and Random Forest. We trained the tree models with categorical cross-entropy (CCE) loss and TabNet with focal loss and CCE, individually. We also applied an early stopping callback monitoring the validation loss and a learning rate scheduler for improving the TabNet performance. The performances of the models were measured by the validation accuracy and F1 score.

## 4. Results and Discussions

### A. Performance comparison

We experimented with two loss functions, CCE loss and focal loss, for training TabNet and trained it for 120 epochs each. Both the models early stopped, and gave their best at epoch 94 with CCE loss and epoch 91 with focal loss while being monitored on the validation loss, Fig. 4. The focal loss was fundamentally designed to handle the imbalanced data. Fig. 5 showcases a comparison between validation metrics when trained on the two loss function and it is inferred that the model with the objective as focal loss performed better than the model with CCE loss. Precisely, there is an uplift of 1.2% in the accuracy.

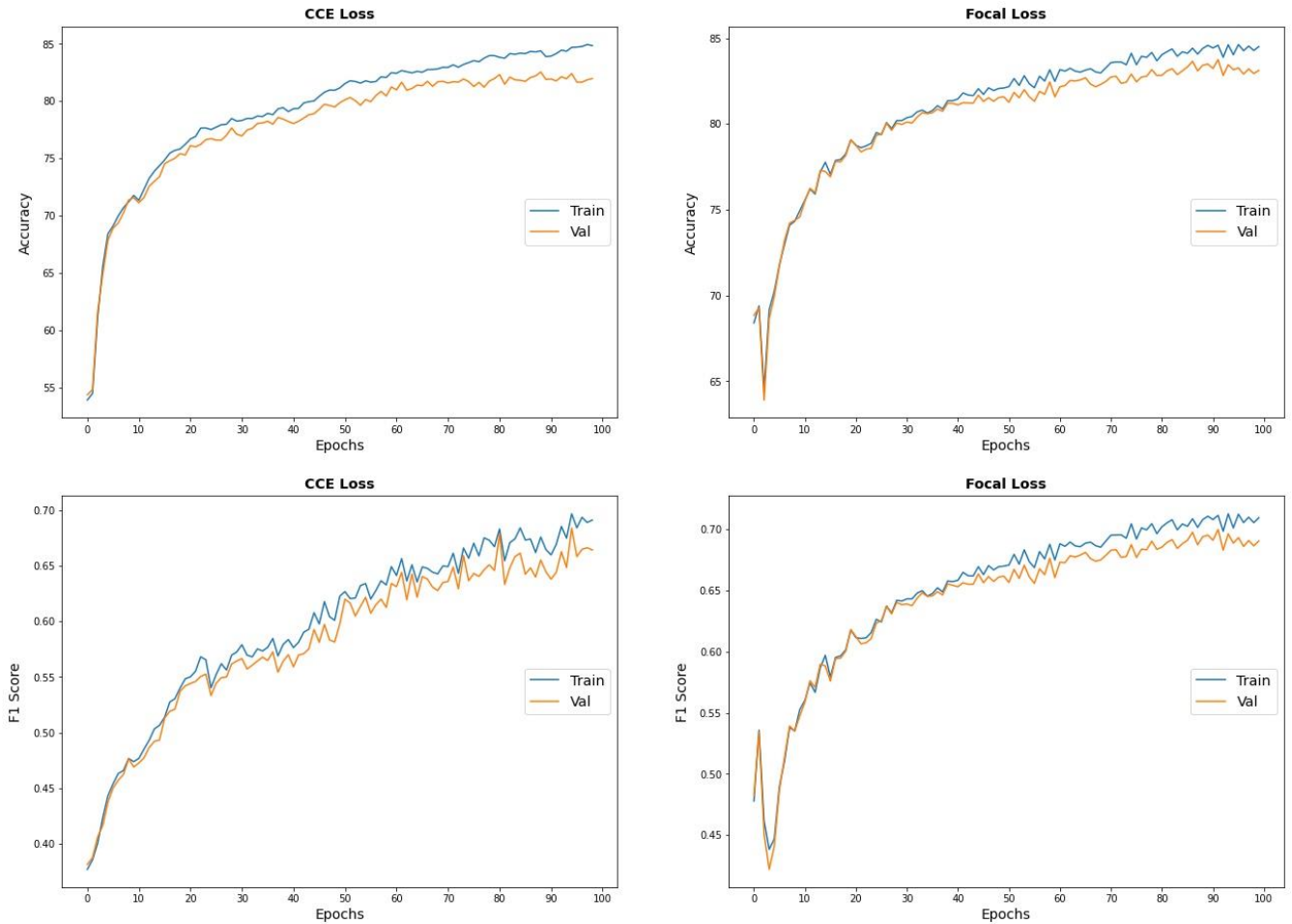

**Figure 4**: Train and validation performance for the best TabNet architecture with different metrics and loss function. (a) [top left] loss function: CCE loss, metrics: accuracy. (b) [top right] loss function: focal loss, metrics: accuracy. (c) [bottom left] loss function: CCE loss, metrics: f1 score. (d) [bottom right] loss function: focal loss, metrics: f1 score.

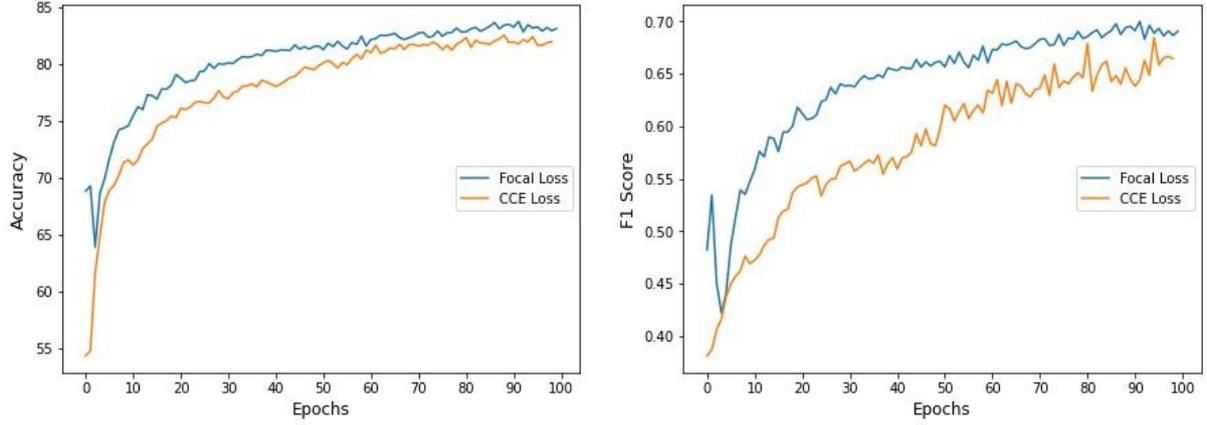

**Figure 5**: Performance comparison using validation metrics (accuracy on left and f1score on right) between focal loss and categorical cross-entropy loss when trained on the same TabNet architecture.

Table 1 shows that TabNet outperforms tree-based models on the water pump dataset. TabNet, due to its DNN capabilities, can easily learn non-linearities in the data, and avoid overfitting, a common problem in neural networks, because of its instance-wise feature selection.

**Table 1: Validation metrics for different models**

| Model | Accuracy (%) | F1 Score |
|---|---|---|
| CatBoost | 75.9 | 0.593 |
| Random Forest | 77.35 | 0.621 |
| LightGBM | 78.68 | 0.648 |
| XGBoost | 79.42 | 0.664 |
| TabNet (CCE loss) | 82.43 | 0.684 |
| **TabNet (focal loss)** | **83.6** | **0.697** |

### B. Model interpretability

The learnable mask implemented in attentive transformer helps in global and instance-wise interpretability, $M[i] \cdot f$ where $f \in \mathbf{R}^{B \times D}$. A traditional neural network lacks these features as the sub-sequent layers are fully connected with the previous layers and there is no mechanism for feature selection at the input level. Table 2 shows the top 5 features selected by the four best performing models.

**Table 2: Feature importance of different models**

| Rank | TabNet | XGBoost | LightGBM | Random Forest |
|---|---|---|---|---|
| 1 | Longitude | Quantity | date_recorded | waterpoint_type |
| 2 | Latitude | waterpoint_type | latitude | quantity_group |
| 3 | wpt_name | extraction_type_group | longitude | quantity |
| 4 | Quantity | region | lga | extraction_type_group |
| 5 | quantity_group | region_code | ward | region |

## 5. Conclusion

We analyzed and applied TabNet, a DNN designed for tabular data, to predict the water pump repair status in wells in Tanzania. The predictions would help in proper planning of the maintenance of the pumps and minimize the water crisis faced due to the failure in its operations. TabNet employs sequential attention to select the salient features at every decision step. The instance-wise feature selection helps in model interpretability. We demonstrated that TabNet outperformed famous gradient boosted tree-based models like XGBoost, LightGBM, CatBoost. Furthermore, TabNet, when trained with a focal loss instead of the categorical cross-entropy loss, showed a further uplift in the performance on the imbalanced dataset.